\documentclass{article}

\usepackage{microtype}
\usepackage{graphicx}
\usepackage{subfigure}
\usepackage{booktabs} % for professional tables

\usepackage{hyperref}

\usepackage[accepted]{icml2024}

\usepackage{amsmath}
\usepackage{amssymb}
\usepackage{mathtools}
\usepackage{amsthm}

\usepackage[capitalize,noabbrev]{cleveref}

\usepackage{url}
\usepackage{wrapfig}
\usepackage{lipsum}
\usepackage{colortbl,xcolor}
\usepackage{multirow}

\usepackage{fontawesome}

\theoremstyle{plain}

\theoremstyle{definition}

\theoremstyle{remark}

\usepackage[textsize=tiny]{todonotes}
\newcommand{\revise}[1]{\textcolor{black}{#1}}

\icmltitlerunning{Split-Ensemble: Efficient OOD-aware Ensemble via Task and Model Splitting}

\begin{document}

\twocolumn[
\icmltitle{Split-Ensemble: Efficient OOD-aware Ensemble via Task and Model Splitting}



\icmlsetsymbol{equal}{*}

\begin{icmlauthorlist}
\icmlauthor{Anthony Chen}{equal,PKU}
\icmlauthor{Huanrui Yang}{equal,UCB}
\icmlauthor{Yulu Gan}{equal,PKU}
\icmlauthor{Denis A Gudovskiy}{PAN}
\icmlauthor{Zhen Dong}{UCB}
\icmlauthor{Haofan Wang}{CMU}
\icmlauthor{Tomoyuki Okuno}{PAN}
\icmlauthor{Yohei Nakata}{PAN}
\icmlauthor{Kurt Keutzer}{UCB}
\icmlauthor{Shanghang Zhang}{PKU}
\end{icmlauthorlist}

\icmlaffiliation{PKU}{School of Computer Science, Peking University}
\icmlaffiliation{UCB}{University of California, Berkeley}
\icmlaffiliation{CMU}{Carnegie Mellon University}
\icmlaffiliation{PAN}{Panasonic Holdings Corporation}

\icmlcorrespondingauthor{Shanghang Zhang}{shanghang@pku.edu.cn}
\icmlcorrespondingauthor{Huanrui Yang}{huanrui@berkeley.edu}

\icmlkeywords{OOD Detection, Ensemble, Efficient deep learning}

\vskip 0.3in
]



\printAffiliationsAndNotice{\icmlEqualContribution} % otherwise use the standard text.

\begin{abstract}
Uncertainty estimation is crucial for deep learning models to detect out-of-distribution (OOD) inputs. However, the naive deep learning classifiers produce uncalibrated uncertainty for OOD data. Improving the uncertainty estimation typically requires external data for OOD-aware training or considerable costs to build an ensemble. 
In this work, we improve on uncertainty estimation without extra OOD data or additional inference costs using an alternative \textit{Split-Ensemble} method. 
Specifically, we propose a novel \textit{subtask-splitting} ensemble training objective where a task is split into several complementary subtasks based on feature similarity. Each subtask considers part of the data as in-distribution while all the rest as OOD data. 
Diverse submodels can therefore be trained on each subtask with OOD-aware objectives, learning generalizable uncertainty estimation.
%Such OOD-aware training can happen without external data because each subtask is performed by a corresponding submodel of the Split-Ensemble. 
To avoid overheads, we enable low-level feature sharing among submodels, building a tree-like Split-Ensemble architecture via iterative splitting and pruning. 
%This leads to improved accuracy and uncertainty estimation across submodels under a fixed ensemble computation budget.
%start from a shared backbone model for all subtasks and perform its splitting and pruning into a tree-like architecture during training using the proposed algorithm. This results in an efficient subtask ensemble, where each subtask is trained for classification with uncertainty estimation. 
Empirical study shows Split-Ensemble, without additional computational cost, improves accuracy over a single model by 0.8\%, 1.8\%, and 25.5\% on CIFAR-10, CIFAR-100, and Tiny-ImageNet, respectively. OOD detection for the same backbone and in-distribution datasets surpasses a single model baseline by 2.2\%, 8.1\%, and 29.6\% in mean AUROC, respectively.
\end{abstract}

\section{Introduction}
\label{sec:intro}

%Start with the importance of uncertainty estimation and OOD detection. Single DNN not good enough (over confidence)
Deep learning models achieve high accuracy metrics when applied to in-distribution (ID) data. However, such models deployed in the real world can also face corrupted, perturbed, or out-of-distribution inputs~\citep{hendrycks2019benchmarking}. Then, model predictions may not be reliable. Therefore, estimation of the epistemic uncertainty with OOD detection is crucial for trustworthy models~\citep{gal2016dropout}.

In general, uncertainty estimation is not a trivial task. Practitioners often consider various statistics derived from the uncalibrated outputs of softmax classifiers as confidence scores~\citep{hendrycks22a}. On the other hand, deep ensembling is another popular approach~\citep{lakshminarayanan2017simple}, where uncertainty can be derived from predictions of independently trained deep networks. However, deep ensembles come with large memory and computational costs, which grow linearly with the ensemble size. Recent research investigates strategies to share and reuse parameters and processing across ensemble submodels~\citep{gal2016dropout, wen2020batchensemble, turkoglu2022film}. Though these techniques reduce memory overheads, they suffer from the reduced submodel diversity and, thus, uncertainty calibration. Moreover, their computational costs remain similar to the naive deep ensemble since the inference through each individual submodel is still required.

More advanced methods for uncertainty estimation using a single model include temperature scaling, adversarial perturbation of inputs~\citep{liang2017enhancing}, and classifiers with the explicit OOD class trained by OOD-aware outlier exposure training~\citep{hendrycks2018deep}. However, as outlier exposure methods typically outperform other approaches that do not utilize external OOD-like data, OOD data distribution can be unknown or unavailable to implement effective outlier exposure training in practical applications.

\begin{figure*}[t]
    \centering
    \includegraphics[width=\linewidth, keepaspectratio=true]{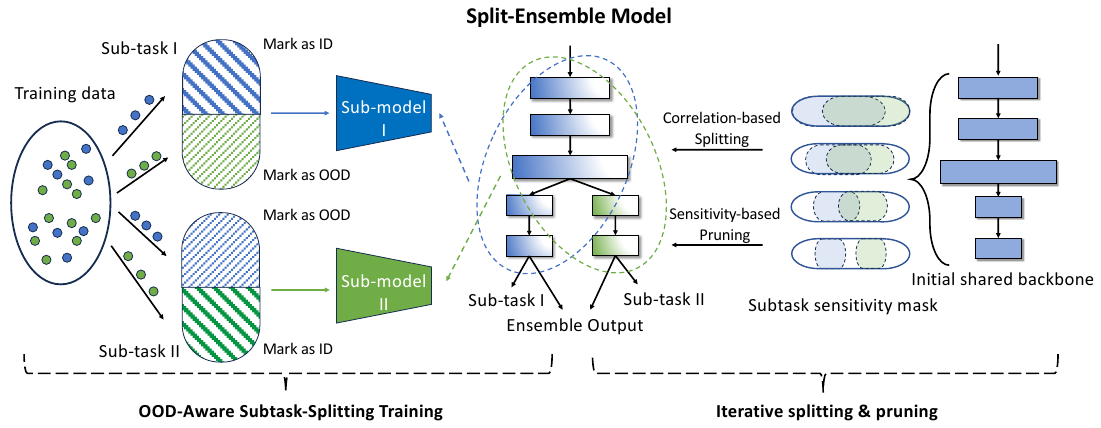}
    \caption{\textbf{Overview.} We split an original task into complementary subtasks to create objectives for submodel training. All submodels form an ensemble to perform the original task, and, importantly, each submodel can be trained with OOD-aware objectives of the subtask as proposed in Section~\ref{sec:split}. To implement an efficient Split-Ensemble architecture, we start with a shared backbone and iteratively perform splitting and pruning based on subtask similarity and sensitivity described in Section~\ref{sec:prune}.}
    \label{fig:intro}
\end{figure*}

This work aims to propose novel training objectives and architectures to build a classifier with a single-model cost, without external OOD proxy data, while achieving ensemble-level performance and uncertainty estimation. 
To avoid the redundancy of having multiple ensemble submodels learn the same task, we, instead, split an original multiclass classification task into multiple complementary subtasks. As illustrated in Figure~\ref{fig:intro}, each subtask is defined by considering a subset of classes in the original training data as ID classes. Then, the rest of the training set is a proxy for OOD distribution for the subtask. Feature similarity is used in grouping the subtasks so that the ID and OOD classes can be well separated. This enables our novel \textit{training objective with subtask splitting}, where each submodel learns an OOD-aware objective for generalizable uncertainty estimation without external data. Finally, an ensemble of all submodels implements the original multiclass classification.

Our splitting objective requires a method to design computationally efficient submodel architectures for each subtask. 
Most subtask processing, as part of the original task, can utilize similar low-level features. Hence, it is possible to share early layers across submodels. Moreover, as each subtask is easier than the original task, we can use lighter architectures for the latter unshared processing in submodels when compared to the backbone design of the original task.
Considering these two key observations, we propose a novel \textit{iterative splitting and pruning} algorithm to learn a tree-like Split-Ensemble model. As illustrated in Figure~\ref{fig:intro}, the Split-Ensemble shares early layers for all submodels. Then, they gradually branch out into different subgroups, resulting in completely independent branches for each submodel towards the last layers. Global structural pruning is further performed on all the branches to remove redundancies in submodels. Given the potential large design space of Split-Ensemble architectures, we propose correlation-based splitting and pruning criteria based on the sensitivity of model weights to each subtask's objective. This method enables automated architecture design through a single training run for our Split-Ensemble.

In summary, the paper makes the following contributions:
\begin{itemize}
    \item We propose a subtask-splitting training objective for OOD-aware ensemble training without external data.
    \item We propose a dynamic splitting and pruning algorithm to build an efficient tree-like Split-Ensemble architecture corresponding to the subtask splitting.
    \item We empirically show that the proposed Split-Ensemble approach significantly improves accuracy and OOD detection over a single model baseline with a similar computational cost, and outperforms larger ensemble baselines by a factor of $4\times$.
\end{itemize}

In the following sections, we first discuss related work in Section~\ref{sec:related}, followed by deriving the subtask-splitting training objectives in Section~\ref{sec:split}, then propose criteria and algorithm for splitting and pruning mechanism in Section~\ref{sec:prune}, and finally present experiment results in Section~\ref{sec:exp}.
\section{Related Work}
\label{sec:related}

\begin{figure*}[!t]
    \centering
    \includegraphics[width=1.\linewidth, keepaspectratio=true]{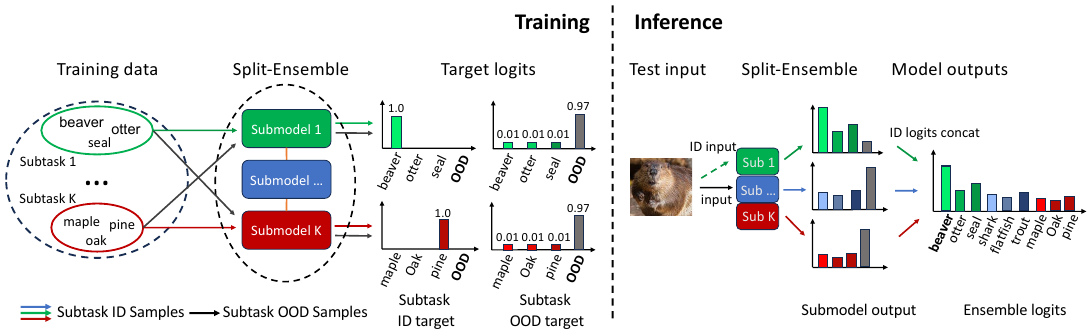 }
    \caption{\textbf{Subtask splitting.} Each submodel learns its subtask using a subset of the original training data. OOD detection by outlier exposure training is realized using other subtasks' examples. Concatenated ID logits from all submodels implement the original multiclass classification task.}
    \label{fig:subtask}
    \vspace{-10pt}
\end{figure*}

\subsection{OOD Detection and OOD-aware Training}

OOD detection has a long history of research, including methods applicable to deep neural networks (DNNs). \citet{4032779} firstly propose to use classification confidence scores to perform in-domain verification through a linear discriminant model. \citet{liang2017enhancing} enhance OOD detection by temperature scaling and adversarial perturbations. \citet{Papadopoulos_2021} propose a method to explicitly make the model aware of OOD using an outlier-exposure training objective with OOD proxy data. \citet{9359272} investigate the differences between OOD-unaware/-aware DNNs in model performance, robustness, and uncertainty. \citet{NEURIPS2020_28e209b6} propose a few-shot learning method for detecting OOD samples. Besides supervised learning, \citet{winkens2020contrastive,sehwag2021ssd} investigate self-supervised OOD detector based on contrastive learning. \citet{pmlr-v162-wang22aq} propose partial and asymmetric supervised contrastive Learning (PASCL) to distinguish between tail-class ID samples and OOD samples. However, the above single-model OOD detectors cannot be implemented without the availability of OOD proxy data. In contrast, our work proposes a novel subtask-splitting training objective to allow OOD-aware learning without external data.

\subsection{Deep Ensembles}

Ensemble methods improve performance and uncertainty estimation by using predictions of multiple members, known as submodels. \citet{lakshminarayanan2017simple} propose the foundation for estimating uncertainty in neural networks using ensemble techniques.
However, computational and memory costs grow linearly with the number of submodels in deep ensembles. To improve efficiency, \citet{mimo2021} replace single-input and single-output layers with multiple-input and multiple-output layers, \citet{gal2016dropout} extract model uncertainty using random dropouts, and \citet{durasov2021masksembles} utilize fixed binary masks to specify network parameters to be dropped. \citet{wen2020batchensemble} enhance efficiency by expanding layer weights using low-rank matrices, \citet{turkoglu2022film} adopt feature-wise linear modulation to instantiate submodels from a shared backbone, and \citet{valdenegro2023sub} ensembles only a selection of layers. These methods aim to mitigate the parameter overheads associated with deep ensembles. However, they cannot reduce computational costs because each submodel runs independently. Split-Ensemble overcomes the redundancy of ensemble processing by having submodels that run complementary subtasks with layer sharing. In addition, we further optimize Split-Ensemble design with tree-like architecture by splitting and pruning.

\subsection{Efficient Multi-task Learning}

With the subtask splitting, our method also falls into the domain of multi-task learning. Given the expense of training individual models for each task, research has been conducted to train a single model for multiple similar tasks.
\citet{sharma2017learning} propose an efficient multi-task learning framework by simultaneously training multiple tasks. \citet{10.1145/3404835.3463022} design Multiple-level Sparse Sharing Model (MSSM), which can learn features selectively with knowledge shared across tasks. \citet{Sun_2021_ICCV} introduce Task Switching Networks (TSNs), a task-conditioned architecture with a single unified encoder/decoder for efficient multi-task learning. \citet{zhang2022automtl} develop AutoMTL that automates efficient MTL model development for vision tasks. \citet{sun2022disparse} propose a pruning algorithm on a shared backbone for multiple tasks.
In this work, we explore correlations between multiple subtasks to design a novel splitting and pruning algorithm.
Previous works on split-based structure search consider layer splitting~\citep{wang2019energy,wu2019splitting,wu2020steepest} to increase the number of filters in certain layers for better model capacity. Split-Ensemble, on the other hand, uses architecture splitting as a way of deriving efficient architecture under a multi-task learning scenario of subtask-splitting training, leading to a novel tree-like architecture.
%\section{Subtask-splitting training}
\section{Subtask Splitting Method}
\label{sec:split}

\begin{figure*}[!t]
    \centering
    \includegraphics[width=.8\linewidth, keepaspectratio=true]{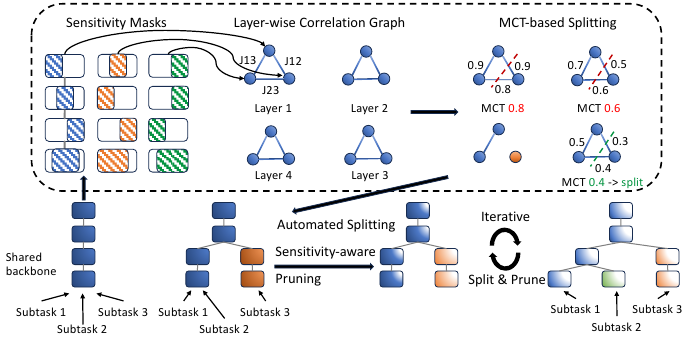}
    \caption{\textbf{Iterative splitting and pruning.} Starting from a shared backbone, we compute the layer-wise sensitivity mask $\mathcal{M}$ for each subtask loss, and calculate pair-wise IoU score $J$ across different subtasks for the layer-wise correlation graph. Model is split at the layer with a small minimal cutting threshold (MCT), and, then, is pruned globally. Applying splitting and pruning in an iterative fashion leads to the final Split-Ensemble architecture that satisfies computational cost constraints.}
    \label{fig:itersplit}
\end{figure*}

In this section, we provide the definition of subtask splitting given a full classification task in Section~\ref{ssec:task_split}. We derive proper training objectives to improve both accuracy and uncertainty calibration for submodels learning these subtasks in Section~\ref{ssec:sub_train}, and how final classification and OOD detection are performed with the ensemble of subtask models in Section~\ref{ssec:ens_train}, as in Figure~\ref{fig:subtask}. %\yulu{Reminder - add sec. ref for a more coherent reading experience.}

\subsection{Complementary Subtask Splitting}
\label{ssec:task_split}

Consider a classification task with $N$ classes in total. Here we define the process of complementary subtask splitting as grouping the classes into $n$ groups with $K_i$ classes in the $i$-th group, where $\sum_{i=1}^n K_i = N$, and each class is only contained in one group. This splits the original classification task into $n$ subtasks with the $i$-th class considering only the $K_i$ classes as in-distribution (ID), while all the other classes are out-of-distribution (OOD) for this subtask.

For training, it is possible to train a submodel for each subtask separately with only the data within the $K_i$ classes. However, at inference, we do not know in advance which submodel we should assign an input with the unknown class to. Therefore, each submodel still needs to handle images from all the classes, not only within the $K_i$ classes. To address that, we add an additional class, namely ``OOD'' class, to the subtask to make it a $(K_i+1)$-way classification task. In the training of the corresponding submodel, we use the entire dataset of the original task with a label conversion. For images in the $K_i$ ID classes of the subtask, we assign them label $0$ through $K_i-1$ in the subtask. For all the other images from the $N-K_i$ OOD classes, we assign the same label $K_i$ to them indicating the OOD class. In this way, a well-trained submodel classifier on the subtask can correctly classify the images within its ID classes, and reject other classes as OOD. Then each input image from the original task will be correctly classified by one and only one submodel, while being rejected by all others.

The grouping strategy for the subtask splitting will affect the final performance. To enable each submodel to learn a generalizable distinction between the ID and OOD classes more easily, we intend to have the ID classes semantically closer to each other while away from the OOD classes. 
In most large-scale real-world datasets, it is by nature that some classes are semantically closer than others, and all classes can be grouped into multiple semantically close groups (e.g. the superclass in CIFAR-100~\citep{krizhevsky2009learning} or the WordNet hierarchy in ImageNet~\citep{deng2009imagenet}). In the case where the semantically similar groups are not naturally available (e.g. in CIFAR-10), we cluster the classes based on mean hidden layer features of each class's training data on a pretrained classifier of the original task, where classes with closer features are grouped together. We empirically verify our intuition on the semantically close grouping in Table~\ref{tab:ablate_group} in the Appendix.

\subsection{Submodel Training Objective}
\label{ssec:sub_train}

Here we derive the training objective for the submodel to learn generalizable uncertainty estimation on each subtask. Without loss of generality, we consider a subtask with $K$ in-distribution classes out of the total $N$ classes in the derivation. The existence of OOD classes in each subtask enables us to perform OOD-aware training for each submodel, without utilizing external data outside of the original task. 
Hence, we take inspiration from outlier exposure~\citep{hendrycks2018deep}, where we train with normal one-hot label for ID data, but use a uniform label for OOD data to prevent the model from over-confidence. Formally, for an input from OOD classes, we set the $i$-th element of the label $\hat{y}^{OOD}$ as
\begin{equation}
    \label{equ:OE_label}
    \hat{y}^{OOD}_i = \begin{cases}
        1/N & 0\leq i \leq K-1 \\
        (N-K)/N & i = K
    \end{cases},
\end{equation}
where class $K$ corresponds to the OOD class.
Note that we set the subtask target of ID classes to $1/N$ instead of $1/K$ to make the ID logits comparable across all submodels with different amounts of classes when facing an OOD input. The OOD logit at class $K$ is therefore used as a ``sink'' to assign the remaining softmax probability. Optimally, all submodels will output a low max probability of $1/N$ when facing an OOD input of the original task, which is as desired for the outlier exposure objective~\citep{hendrycks2018deep}. 
%We substitute the OOD class target in Equation~(\ref{equ:OE_label}) into the loss formulation of Equation~(\ref{equ:CB_loss}) to get $\mathcal{L}_{CB}(X,\hat{Y})$, which we use to train each submodel.

Under the setting of our subtask, each ID class only consists the data from one class in the original task, yet the OOD class corresponds to all data in the rest $N-K$ classes, leading to a $(N-K)\times$ training data to other classes. Directly training with the imbalanced class will lead to significant bias over the OOD class.
We refer to recent advances in imbalanced training and use the class-balance reweighing~\citep{cui2019class} as the loss of each class. The weight $w_i$ of class $i$ of the subtask is formulated as
\begin{equation}
    \label{equ:CB_weight}
    w_i = \begin{cases}
        \frac{1-\beta}{1-\beta^n} & 0\leq i \leq K-1 \\
        \frac{1-\beta}{1-\beta^{(N-K)n}} & i = K
    \end{cases},
\end{equation}
where $n$ is the amount of data in each class of the original task, and $\beta\in[0,1)$ is a hyperparameter balancing the weight. We follow~\citet{cui2019class} to set $\beta=0.9999$, where ablation study shows the choice of $\beta$ does not affect the final performance metric. We apply the reweighing on the binary cross entropy (BCE) loss to formulate the submodel training objective $\mathcal{L}_{CB}(X,Y)$ with submodel output logits $X$ and label $Y$ as
\begin{equation}
\small
    \label{equ:CB_loss}
    \mathcal{L}_{CB} = \mathbb{E}_{X} \sum_{i=0}^K \Bigl[-w_i\bigl(\hat{y}_i \log\sigma(x_i) + (1-\hat{y}_i)\log(1-\sigma(x_i)) \bigr)\Bigr], 
\end{equation}
where $\sigma(\cdot)$ denotes the Sigmoid function, $x_i$ is the $i$-th element of a output logit $x$, and $\hat{y}_i$ follows~\cref{equ:OE_label}. Note that although the submodel learns a multi-class classification as illustrated in~\cref{fig:subtask}, we use the summation of BCE loss of each individual class from $0$ through $k$ as the submodel training objective. This choice of assuming each category is classified independently has been shown to have advantages in learning from imbalanced datasets~\citep{cui2019class}. It also gives us more flexibility in controlling the target labels and class-wise reweighting.

\subsection{Ensemble Training and Inference}
\label{ssec:ens_train}

To get the final output logits for the original task, we concatenate the ID class logits from each submodel into a $N$-dimensional vector. Then the classification can be performed with an argmax of the concatenated logits. In order to calibrate the range of logits across all submodels, we perform a joint training of all submodels, with the objective
\begin{equation}
    \label{equ:ENS_loss}
    \mathcal{L}_{ens} = \sum_i \mathcal{L}_{CB}^i(X_i,\hat{Y}_i) + \lambda \mathcal{L}_{CE}(X,Y).
\end{equation}

Here $X_i$ and $\hat{Y}_i$ denote the output logits and the subtask target of submodel $i$, as formulated in Section~\ref{ssec:sub_train}. $X$ denotes the concatenated ID logits, $Y$ denotes the label of the original task, and $\mathcal{L}_{CE}$ is the cross entropy loss. Hyperparameter $\lambda$ balances the losses. Empirically, we find that a small $\lambda$ (e.g. 1e-4) is enough for the logits ranges to calibrate across submodels, while not driving the ID logits of each submodel to be overconfident. 

For uncertainty estimation, we compute the probability of the ensemble model outputting a label $y$ from the contribution of submodel $f_i$ given an input $z$ as $p(y|z) = p(y|z,f_i) \times p(f_i|z)$.
Here, $p(y|z,f_i)$ can be estimated with the softmax probability of the $y$ class at the output of submodel $f_i$. $p(f_i|z)$ can be estimated with $1-$ the softmax probability of the OOD class of submodel $f_i$, as the probability that $f_i$ provides a valid ID output for the ensemble.
With the design of the OOD-aware training objective in Equation~(\ref{equ:OE_label}), we use $p(y|z)$ as the OOD detection criteria. Specifically, a single threshold will be selected so that all input with a smaller probability than the threshold will be considered as OOD sample. 
%Besides directly using the submodel softmax probability in estimating $p(y|z)$, our model can also be combined with temperature scaling and input perturbation as proposed in ODIN~\citep{liang2017enhancing} to improve the OOD detection.

%\section{Architecture splitting and pruning}
\section{Architecture with Splitting and Pruning}
\label{sec:prune}

In this section, we propose the Split-Ensemble architecture to effectively and efficiently learn the aforementioned subtask-splitting training task. Starting from a single backbone, we discuss the process of submodel splitting in Section~\ref{ssec:arch_split} and formulate the criteria to prune unimportant structures in Section~\ref{ssec:arch_prune}. Figure~\ref{fig:itersplit} overviews our pipeline.

\subsection{Correlation-based Automated Splitting}
\label{ssec:arch_split}

Intuitively, with the subtasks split from the original task, submodels should be able to share the low-level features learned in the early layers of the model, while using diverse high-level filters in later layers in each subtask. 
Given the difficulty of merging independently training submodels during training, we instead start our ensemble training from a single backbone model for the original task. We consider all the submodels using the same architecture with the same parameters of the backbone model, only with independent fully-connected (FC) classifiers at the end. 

The question therefore becomes: \textbf{On which layer shall we split a submodel from the shared backbone?} Here we propose an automated splitting strategy that can learn a proper splitting architecture given the split subtasks. Intuitively, two submodels can share the same layer if both of their tasks benefit from the existing weights (i.e., removing a filter will hurt both subtasks simultaneously). Otherwise, if the weight sensitive to subtask 1 can be removed for subtask 2 without harm, and vice versa, then it would be worth splitting the architecture and shrinking the submodels separately. 
Formally, for submodels sharing the same layer with weight $W$, we perform a single-shot sensitivity estimation of all the weight elements $w_j$ in $W$ on the loss $\mathcal{L}_{CB}^i$ of each subtask $i$ respectively as
\begin{equation}
    \label{equ:SNIP}
    s_j^i = \frac{|g(w_j)|}{\sum_{w_k\in W}|g(w_k)|},\ g(w_j) = w_j \nabla_{w_j} \mathcal{L}_{CB}^i(W),
\end{equation}
following the criteria proposed in SNIP~\citep{lee2018snip}. Then we perform a Top-K masking to select the $K$ weight elements with the largest sensitivity, forming a sensitive mask $\mathcal{M}_i$ for submodel $i$.
The weight element lies in the intersection of the two masks $\mathcal{M}_i \cap \mathcal{M}_j$ are sensitive for both subtasks, while other elements in the union $\mathcal{M}_i \cup \mathcal{M}_j$ but not in the intersection is only sensitive to one subtask. We use the Intersection over Union (IoU) score to measure the pair-wise mask correlation as $J_{ij} = \frac{|\mathcal{M}_i \cap \mathcal{M}_j|}{|\mathcal{M}_i \cup \mathcal{M}_j|}$, where $|\cdot|$ denotes the cardinality of a set. It has been observed in previous multi-task pruning work that pruning mask correlations will be high in the early layers and drop sharply towards later layers~\citep{sun2022disparse}, as later layers learn more high-level features that are diverse across subtasks. A pair of submodels can be split at the earliest layer where the IoU score drops below a pre-defined threshold. The architecture and parameters of the new branch are initialized as the exact copy of the original layers it is splitting from, which guarantees the same model functionality before and after the split. The branches will then be updated and pruned independently according to their subtask objectives after the splitting is performed.

In the case of multiple submodels, we compute the pairwise IoU score between each pair of submodels and build a ``\textit{correlation graph}'' for each layer. The correlation graph is constructed as a weighted complete graph $C = (V,E)$ with each submodel being a node $v\in V$, and the IoU score between two submodels $J_{uv}$ is assigned as the weight of the edge $(u,v)\in E$ between the corresponding nodes. Then a split of the model is equivalent to performing a cut on the correlation graph to form two separated subgraphs $S$ and $T$. 
Here we propose a measurement of ``\textit{Minimal Cutting Threshold}'' (MCT), which is the minimal correlation threshold for edge removal that can cut the correlation graph into two. Formally, the MCT of a correlation graph $C$ is defined as
\begin{equation}
    \label{equ:MCT}
    \small
    MCT(C) = \min_{S,T} \left[ \max_{u\in S, v\in T, (u,v)\in E} J_{uv} \right],\ s.t.\ S+T=V .
\end{equation}

A small MCT indicates that a group of submodels has a weak correlation with the rest, therefore they can be separated from the shared architecture. In practice, we will iteratively split the earliest layer with an MCT lower than a predefined value in a branch shared by multiple submodels, until all submodels have individual branches or the training ends.
The splitting strategy will turn a single backbone model into a tree-like architecture, as illustrated in Figure~\ref{fig:itersplit}. 

\subsection{Sensitivity-aware Global Pruning}
\label{ssec:arch_prune}

To remove the redundancy in the submodels for simpler subtasks, we perform global structural pruning on the Split-Ensemble architecture. We perform structural sensitivity estimation on a group of weight element $w_s$ belonging to a structure $\mathcal{S}$ for the loss $\mathcal{L}_{CB}^i$ of each subtask $i$. We utilize the Hessian importance estimation~\citep{yang2023global}, which is computed as
\begin{equation}
    \label{equ:imp}
    \mathcal{I}^i(\mathcal{S}) = \left(\sum_{s\in\mathcal{S}}w_s \nabla_{w_s}\mathcal{L}_{CB}^i(w_s)\right)^2.
\end{equation}

It has been shown that this importance score is comparable across different layers and different types of structural components~\citep{yang2023global}, making it a good candidate for global pruning. Then, we greedily prune a fixed number of filters with the smallest $\mathcal{I}^i$ in submodel $i$. 
In the case where multiple submodels are sharing the structure, we separately rank the importance of this structure in each submodel, and will only prune the filters that are prunable for all the submodels sharing it.

%\subsection{Iterative splitting and pruning}
%\label{ssec:arch_design}
Putting everything together, we iteratively perform the aforementioned automated splitting and pruning process during the training of the Split-Ensemble model. Splitting and pruning are performed alternatively. Removing commonly unimportant structures will reduce the sensitivity correlation in the remaining parameters, enabling further splitting of the submodels. In the meantime, having a new split enables additional model capacity for further pruning. The splitting will be fixed when all submodels have an individual branch towards later layers of the model. Pruning will be stopped when the Floating-point Operations (FLOPs) of the Split-Ensemble architecture meet a predefined computational budget, typically the FLOPs of the original backbone model. The Split-Ensemble model will then train with the fixed model architecture for the remaining training epochs.
The detailed process of Split-Ensemble training is provided in the pseudo-code in Algorithm~\ref{alg} of Appendix~\ref{ap:alg}.

\begin{table*}[!t]
\caption{\textbf{OOD detection results}. Models trained on ID dataset are evaluated against multiple OOD datasets. The results are reported for models with ResNet-18 backbone. \revise{FPR and detection error are evaluated with the threshold achieving 95\% TPR.}}
\label{tab:ood-results}
\vspace{0.1cm}
\centering
\resizebox{\linewidth}{!}{
\begin{tabular}{llcccc}
\toprule
{\bf ID} & {\bf OOD}  &\bf{FPR} & \bf{Det. Error} & {\bf AUROC} & {\bf AUPR} \\
{\bf dataset} & {\bf dataset} &\bf{(95\% TPR)}$\downarrow$ & \bf{\revise{(95\% TPR)}} $\downarrow$& $\uparrow$ & $\uparrow$ \\
\midrule
\multirow{8}{0.12\linewidth}{CIFAR-10}  
&   & \multicolumn{4}{c}{{ \bf{Single Model / Naive Ensemble (4x) / Split-Ensemble (ours)}}} \\
\cmidrule{3-6}

& CIFAR-100 & 56.9 / \revise{50.6} / \textbf{47.9} & 30.9 / \revise{27.8} / \textbf{26.4} & 87.4 / \revise{87.8} / \textbf{89.6} & 85.7 / \revise{85.6} / \textbf{89.5} \\
& TinyImageNet (crop) & 30.9 / \textbf{\revise{29.9}} / 39.2 & 17.9 / \textbf{\revise{17.4}} / 22.1 & 93.1 / \revise{94.4} / \textbf{94.9} & 96.0 / \revise{93.8} / \textbf{96.4}  \\ 
& TinyImageNet (resize) & 54.9 / \revise{50.3} / \textbf{46.0} & 29.9 / \revise{27.7} / \textbf{25.5} & 87.5 / \revise{89.3} / \textbf{91.7} & 86.2 / \revise{88.0} / \textbf{92.8}\\ 
& SVHN & 48.4 / \revise{31.1} / \textbf{30.5} & 17.0 / \revise{12.2} / \textbf{12.1} & 91.9 / \revise{93.8} / \textbf{95.2} & 84.0 / \revise{85.8} / \textbf{91.9} \\
& LSUN (crop) & 27.5 / \textbf{\revise{18.7}} / 37.5 & 16.3 / \textbf{\revise{11.9}} / 21.3 & 92.1 / \textbf{\revise{95.9}} / 95.3 & \textbf{96.8} / \revise{94.8} / \textbf{96.8}\\ 
& LSUN (resize) & 49.4 / \revise{34.6} / \textbf{33.2} & 27.2 / \revise{19.8} / \textbf{19.1} & 90.5 / \revise{93.4} / \textbf{94.5} & 90.7 / \revise{93.3} / \textbf{95.7}\\ 
& Uniform  & 83.0 / \revise{85.3} / \textbf{63.7} & 76.3 / \revise{78.1} / \textbf{58.7} & 91.9 / \revise{88.5} / \textbf{92.5} & 99.2 / \revise{98.8} / \textbf{99.3}\\ 
& Gaussian  & \textbf{9.4} / \revise{95.4} / 33.0 & \textbf{9.3} / \revise{87.2} / 30.5 & \textbf{97.7} / \revise{85.6} / 95.7 & \textbf{99.8} / \revise{98.3} / 99.6  \\ 
& \cellcolor{gray!10}\textbf{Mean} & \cellcolor{gray!10}45.1 / \cellcolor{gray!10}\revise{49.5} / \textbf{41.4} & \cellcolor{gray!10}28.1 / \cellcolor{gray!10}\revise{35.3} / \textbf{27.0} & \cellcolor{gray!10} 91.5 / \cellcolor{gray!10}\revise{91.1} / \textbf{93.7} &\cellcolor{gray!10} 92.3 / \cellcolor{gray!10}\revise{92.3} / \textbf{95.3}  \\

 %&  All &  & \\
\midrule
\multirow{7}{0.12\linewidth}{CIFAR-100}
& CIFAR-10 & \textbf{76.2} / \revise{78.6} / 78.5 & \textbf{40.6} / \revise{41.8} / 41.7 & \textbf{80.5} / \revise{80.3} / 79.2 & \textbf{83.2} / \revise{82.2} / 81.7\\
& TinyImageNet (crop) & 66.1 / \revise{77.5} / \textbf{58.1} & 41.2 / \revise{49.7} / \textbf{31.6} & 85.8 / \revise{80.3} / \textbf{88.4} & 88.3 / \revise{82.2} / \textbf{90.0} \\ %1000, 0.002
& TinyImageNet (resize)  & \textbf{68.2} / \revise{78.6} / 72.1 & 38.9 / \revise{41.8} / \textbf{38.6} & \textbf{84.4} / \revise{77.5} / 82.7& \textbf{86.9}/ \revise{79.3} / 84.6\\ %1000,	 0.0022
& SVHN & \textbf{60.6} / \revise{75.2} / 75.0 & \textbf{20.4} / \revise{24.5} / 24.4 & \textbf{87.7} / \revise{83.3} / 81.2& \textbf{81.1} / \revise{74.4} / 69.9 \\
& LSUN (crop) & 70.9 / \revise{84.7} / \textbf{64.7} & 44.8 / \revise{49.3} / \textbf{34.9} & 76.7 / \revise{76.7} / \textbf{85.3} & \textbf{86.6} / \revise{79.9} / \textbf{86.6} \\ %1000, 	0.0038,
& LSUN (resize) & \textbf{66.7} / \revise{79.1} / 72.0 & \textbf{35.9} / \revise{42.1} / 38.5 & \textbf{85.4} / \revise{78.3} / 83.2& \textbf{87.9} / \revise{80.1} / 85.6 \\ %1000,	0.0018
%& iSUN &84.8/49.5 & 44.7/27.2& 69.9/90.1& 71.9/91.1&67.0/88.9 \\  %1000, 	0.002
& Uniform & 100.0 / \revise{100.0} / \textbf{95.0} & 90.9 / \revise{90.9} / \textbf{65.9} & 59.2 / \revise{69.1} / \textbf{88.3}& 95.2 / \revise{91.6} / \textbf{98.8} \\   % 1,	0.0024
& Gaussian & 100.0 / \revise{100.0} / \textbf{99.6} & 90.9 / \textbf{\revise{72.5}} / 90.9 & 40.6 / \revise{59.2} / \textbf{63.1} & 92.0 / \revise{95.2} /\textbf{95.5} \\ %1, 	0.0028
& \cellcolor{gray!10}\textbf{Mean} &\cellcolor{gray!10} 76.1 / \revise{84.2} / \textbf{74.0} &\cellcolor{gray!10} 48.9 / \revise{52.3} / \textbf{45.7} &\cellcolor{gray!10} 73.9 / \revise{75.6} / \textbf{82.0} & \cellcolor{gray!10}\textbf{87.3} / \revise{83.7} / 86.6  \\
\midrule
\multirow{7}{0.12\linewidth}{Tiny-IMNET} 
& CIFAR-10 & 99.3 / \textbf{97.7} / 100.0 & \textbf{33.3} / 50.3 / \textbf{33.3} & 56.5 / 46.7 / \textbf{81.2} & 48.9 / 48.6 / \textbf{82.7} \\
& CIFAR-100 & 99.2 / \textbf{97.5} / 100.0 & 33.3 / 50.3 / \textbf{9.1} & 54.6 / 46.1 / \textbf{72.6} & 45.5 / 47.5 / \textbf{51.9}  \\
& SVHN & \textbf{95.2} / 97.5 / 100.0 & \textbf{16.1} / 20.1 / \textbf{16.1} & 64.8 / 46.5 / \textbf{83.6} & 38.1 / 26.6 / \textbf{80.2} \\
& LSUN (crop) & 100.0 / 97.5 / \textbf{94.0} & \textbf{33.3} / 50.3 / \textbf{33.3} & 28.9 / 45.9 / \textbf{80.2} & 25.9 / 48.8 / \textbf{78.5}  \\ %1000, 	0.0038,
& LSUN (resize) & 99.8 / \textbf{97.8} / 100.0 & 50.3 / 50.3 / \textbf{33.3} & 44.9 / 45.9 / \textbf{76.3} & 36.5 / 47.4 / \textbf{77.2} \\ %1000,	0.0018
%& iSUN &84.8/49.5 & 44.7/27.2& 69.9/90.1& 71.9/91.1&67.0/88.9 \\  %1000, 	0.002
& Uniform & 100.0 / \textbf{90.2} / 100.0 & 83.3 / \textbf{73.5} / 83.3 & 24.2 / 43.9 / \textbf{63.8} & \textbf{77.7} / 90.2 / \textbf{92.5} \\   % 1,	0.0024
& Gaussian & 100.0 / \textbf{96.7} / 100.0 & 83.3 / \textbf{73.5} / 83.3 & 25.4 / 43.8 / \textbf{49.3}& 78.1 / \textbf{89.9} / 88.1\\ %1, 	0.0028
&\cellcolor{gray!10} \textbf{Mean} & \cellcolor{gray!10}\textbf{99.1} / 96.4 / \textbf{99.1} &\cellcolor{gray!10} \textbf{45.1} / 52.6 / 46.4 &\cellcolor{gray!10} 42.8 / 45.8 / \textbf{72.4} &\cellcolor{gray!10} 50.1 / 57.0 / \textbf{78.7}  \\ 
\bottomrule
\end{tabular}
}
\end{table*}

\begin{table*}[ht]
 \caption{Comparison between previous state-of-the-art ensemble-based methods and ours on the \textbf{SC-OOD CIFAR10-LT benchmarks.}  The results are reported for models with ResNet-18 backbone. Best score in \textbf{bold}, second best \underline{underlined}.}
  \label{tab:cifar10LT-SCOOD}
  \vspace{0.1cm}
  \small
  \centering
  \begin{tabular}{l|c|cccc} 
    \toprule
    Method                & Accuracy $\uparrow$ & ECE $\downarrow$ & FPR95 $\downarrow$ & AUROC $\uparrow$ & AUPR $\uparrow$ \\
    \midrule
     Naive Ensemble        & 12.7        & 50.2               & 98.4                    & 45.3                    & 50.9                   \\
MC-Dropout            & 63.4             & 25.8          & 90.6                    & 66.6                    & 66.1                   \\
MIMO                  & 35.7             & 28.8          & 96.3                    & 55.1                    & 56.9                   \\
MaskEnsemble          & 67.7             & 24.6          & 89.0                    & 66.82                   & 67.4                   \\
BatchEnsemble         & 70.1             & \underline{21.1}          & 87.45                   & 68.0                    & 68.7                   \\
FilmEnsemble          & \underline{72.5} & 21.3                    & \underline{84.32}                 & \underline{75.5}                  & \underline{76.0}                 \\
\midrule
Split-Ensemble (ours) & \textbf{73.7 }   & \textbf{16.5}               & \textbf{80.5}                & \textbf{81.7}                & \textbf{77.6 }    \\
    \bottomrule
  \end{tabular}
\end{table*}

\begin{table*} [!t]
  \caption{\textbf{Classification results on CIFAR-10 and CIFAR-100}. Best score for each metric in \textbf{bold}, second-best \underline{underlined}. We implement all baselines using default hyperparameters. All accuracies are given in percentage with ResNet-18/ResNet-34 as backbone}
  \vspace{0.1cm}
  \label{tab:acc}
  \centering
  \small
  \resizebox{.6\linewidth}{!}{
  \begin{tabular}{l|c|cc}
    \toprule
    %\multicolumn{2}{c}{Part}                   \\
   % \cmidrule(r){1-2}
    Method    & FLOPs & CIFAR-10 Acc ($\uparrow$) & CIFAR-100 Acc ($\uparrow$)  \\
    \midrule
    Single Model      & 1x    & 94.7 / 95.2   & 75.9 / 77.3  \\
    Naive Ensemble & 4x    & \revise{\textbf{95.7}} / \revise{\underline{95.5}}   & \revise{\textbf{80.1}} / \revise{\textbf{80.4}}  \\ 
    \midrule
    MC-Dropout    & 4x    & \revise{93.3} / \revise{90.1}   & \revise{73.3} / \revise{66.3} \\  
    MIMO          & 4x    & \revise{86.8} / \revise{87.5}   & \revise{54.9} / \revise{54.6}  \\
    MaskEnsemble   & 4x    & \revise{94.3} / \revise{90.8}   & \revise{76.0} / \revise{64.8}  \\
    BatchEnsemble  & 4x    & \revise{94.0} / \revise{91.0}   & \revise{75.5} / \revise{66.1}   \\
    FilmEnsemble   & 4x    & \revise{87.8} / \revise{94.3}   & \revise{77.4} / \revise{77.2}   \\
   \midrule
    \textbf{Split-Ensemble (Ours)} & 1x     & \underline{95.5} / \textbf{95.6}   & \underline{77.7} / \underline{77.4} \\
    % \textbf{Split-Ensemble (Ours)} & 4x & 78.27  & 0.135   & \underline{?}  & \underline{?} \\
    \bottomrule 
  \end{tabular}
  }
\end{table*}
%newpage
\section{Experiments}
\label{sec:exp}

In this section, we compare the OOD detection performance and the accuracy of Split-Ensemble against baseline single model and ensemble methods on various datasets in Section~\ref{ssec:OOD} and~\ref{ssec:accuracy} respectively. We provide ablation studies on the design choices of using OOD-aware target in subtask training and selecting MCT threshold for iterative splitting in Section~\ref{ssec:ablation}. Detailed experiment settings are available in Appendix~\ref{sec:implemen}, and additional results in Appendix~\ref{ap:visualize}.

\subsection{Performance on OOD Detection}
\label{ssec:OOD}

As we focus the design of Split-Ensemble on better OOD-aware training, here we compare the OOD detection performance of our method against a single model and naive ensemble baselines with ResNet-18 backbone. All models are trained using the same code under the same settings. 
Table~\ref{tab:ood-results} shows the comparison between the OOD detection performance. We can clearly see that our method outperforms a single model baseline and a 4$\times$ larger naive ensemble across all benchmarks. This improvement shows our OOD-aware training performed on each subtask can generalize to unseen OOD data, without using additional training data. 
We also evaluated our method on CIFAR10-LT with SCOOD benchmark, a more challenging long-tailed dataset. As shown in Table~\ref{tab:cifar10LT-SCOOD}, our approach consistently excels across accuracy, expected calibration error (ECE), and all OOD detection metrics. We further provide additional results on the SC-OOD dataset and against other OOD detection methods in Table~\ref{tab:single-cifar10-SCOOD} and Table~\ref{tab:ensemble-cifar10-SCOOD} in Appendix~\ref{ap:visualize} due to space constraints.

% For OOD detection score computation, we use the max softmax probability for the single model, max softmax probability of the mean logits for the naive ensemble, and use the probability score proposed in Section~\ref{ssec:ens_train} for our Split-Ensemble. A single threshold is used to detect OOD with score lower than the threshold.

% Additionally, we assess our method on CIFAR10-LT, a complex long-tailed dataset, to evaluate its robustness. As evidenced in Table~\ref{tab:cifar10LT-SCOOD}, our approach consistently outperforms in all four metrics.

\subsection{Performance on Classification Accuracy}
\label{ssec:accuracy}
% todo 要不要修改结构
We train Split-Ensemble on CIFAR-10, CIFAR-100 datasets and evaluate its classification accuracy. The results are compared with baseline single model, naive ensemble with 4x submodels, and other parameter-efficient ensemble methods in Table~\ref{tab:acc}. On CIFAR-100, we notice that the Naive Ensemble with independent submodels achieves the best accuracy. Other efficient ensembles cannot reach the same level of accuracy with shared parameters, yet still require the same amount of computation. Our Split-Ensemble, on the other hand, beats not only single model but also other efficient ensemble methods, without additional computational cost.

Additional results on Tiny-ImageNet and ImageNet1K from Table~\ref{tab:imagenet} show that Split-Ensemble can bring consistent performance gain over single model, especially for difficult tasks like ImageNet, where a single model cannot learn well. The improved performance comes from our novel task-splitting training objective, where each submodel can learn faster and better on simpler subtasks, leading to better convergence. The iterative splitting and pruning process further provides efficient architecture for the Split-Ensemble to achieve high performance without computational overhead. 

As observed by~\cite{ovadia2019can}, the performance and uncertainty estimation ability of ensemble methods may be challenged by corruptions in the data. To this end, we compare Split-Ensemble against other baseline methods on the corrupted CIFAR-10-C dataset, and show its robustness under corrupted inputs. The results are presented in~\cref{tab:rob} in~\cref{ap:visualize}.

\begin{table}[tb]
 \caption{\textbf{Classification results on TinyImageNet and ImageNet1K.} Top-1 accuracies (\%) are reported for models with ResNet-18 backbone. Best score in \textbf{bold}.}
  \label{tab:imagenet}
  \vspace{0.1cm}
 \centering
  \small
\begin{tabular}{l|c|c}
\toprule
Method         & TinyImageNet & ImageNet1K  \\
\midrule
Single Model       & 26.1 & 69.0 \\
Naive Ensemble & 44.6 & 69.4 \\
\midrule
Split-Ensemble (ours) & \textbf{51.6} & \textbf{70.9} \\  
\bottomrule
\end{tabular}
\end{table}

\subsection{Ablation Studies}
\label{ssec:ablation}

In this section, we provide the results of exploring the use of OOD-aware target (Equation~(\ref{equ:OE_label})) and the impact of MCT threshold in automated splitting (Section~\ref{ssec:arch_split}). Due to space limitation, we put additional results on ablating the influence of the number of subtask splittings and the grouping of classes for each subtask in Appendix~\ref{ap:visualize}.

\begin{table}[tb]
 \caption{\textbf{Ablation on OOD-aware subtask training.} Models are trained on CIFAR-100. OOD detection is against the CIFAR-10 dataset.}
  \label{tab:ablate_oe}
  \vspace{0.1cm}
  \centering
  \renewcommand\arraystretch{1.2}
  \setlength\tabcolsep{5pt}
  \small
  \resizebox{0.8\linewidth}{!}{
  \begin{tabular}{c|c|cc} 
    \toprule
    \# splits & OOD class target & Accuracy & AUROC \\
    \midrule
     \multirow{2}{0.01\textwidth}{2} & One-hot & 71.0 & 76.0 \\
     & OOD-aware & \textbf{77.7} & \textbf{78.1} \\
     \midrule
     \multirow{2}{0.01\textwidth}{4} & One-hot  & 77.2 & 77.5 \\ 
     &  OOD-aware  & \textbf{78.0} & \textbf{78.2} \\
     \midrule
     \multirow{2}{0.01\textwidth}{5} & One-hot & 77.7 & 77.3 \\ 
     & OOD-aware & \textbf{77.9} & \textbf{78.1} \\
    \bottomrule
  \end{tabular}
  }
\end{table}

\paragraph{OOD-aware target}

In Section~\ref{ssec:sub_train}, we propose to use an outlier exposure-inspired target for the inputs belonging to the OOD class, so as to better calibrate the confidence during submodel training. Table~\ref{tab:ablate_oe} compares the results of training Split-Ensemble submodels with the original one-hot labels for OOD class vs. the proposed OOD-aware targets.
No matter how many subtask splittings we use, using the OOD-aware target significantly improved the AUROC for OOD detection, while also helping the ID accuracy of the model. The results indicate that having an explicit OOD class is inadequate for the submodel to learn generalizable OOD detection, and the OOD-aware training objective is effective. Improving submodel OOD detection also helps ensemble accuracy as the submodels can better distinguish their ID classes from others.

\begin{table}[tb]
 \caption{\textbf{Ablation on MCT thresholds.} Models are trained on CIFAR-100 with 5 subtask splits. OOD detection is against CIFAR-10. Detailed split architectures are visualized in Appendix~\ref{ap:visualize}.}
  \label{tab:ablate_mct}
  \vspace{0.1cm}
  \centering
  \renewcommand\arraystretch{1.2}
  \setlength\tabcolsep{5pt}
  \small
  \resizebox{0.9\linewidth}{!}{
  \begin{tabular}{c|ccccc} 
    \toprule
     MCT threshold & 0.0 (all-share) & 0.1 & 0.2 & 0.4 & 0.7  \\
    \midrule
    Accuracy & 76.2 & 77.9 & \textbf{78.4} & 77.9 & 77.9  \\
    AUROC & 76.7 & 78.0 & 78.8 & \textbf{79.9} & 78.9  \\
    \bottomrule
  \end{tabular}
  }
  \vspace{-10pt}
\end{table}

\paragraph{Automated splitting threshold}

In Section~\ref{ssec:arch_split}, we design our automatic splitting strategy as splitting a (group of) submodels from the backbone when the MCT at a certain layer drops below a predefined threshold. The choice of this MCT threshold is therefore impactful on the final architecture and performance of the Split-Ensemble model. Table~\ref{tab:ablate_mct} explores the model performance as we increase the MCT threshold from 0.0 (all-share). As the threshold increases, the models can branch out easier in earlier layers (see architectures in Figure~\ref{fig:vis_arch} in Appendix~\ref{ap:visualize}), which improves the flexibility for the submodels to learn diverse features for the OOD-aware subtasks, leading to improvements in both ID accuracy and OOD detection. However, more and earlier branches require the use of aggressive pruning to maintain the ensemble under cost constraints, which eventually hurts the model performance. A threshold around 0.4 gives a good balance with adequate diversity (as deep ensemble) and high efficiency (as single model) to the final Split-Ensemble model, leading to good performance.
\section{Conclusions}
\label{sec:con}

In this paper, we introduced the Split-Ensemble method, a new approach to improve single-model accuracy and OOD detection without additional training data or computational overhead. By dividing the learning task into complementary subtasks, we enabled OOD-aware learning without external data. Our split-and-prune algorithm efficiently crafted a tree-like model architecture for the subtasks, balancing performance and computational demands. Empirical results validated the effectiveness of our Split-Ensemble. We hope this work opens up a promising direction for enhancing real-world deep learning applications with task and model splitting, where subtasks and submodel architectures can be co-designed to learn better-calibrated efficient models on complicated tasks.

\section*{Acknowledgement}
We thank Berkeley Deep Drive and Panasonic for supporting this research.

\section*{Impact Statement}
This paper presents work whose goal is to advance the field of Machine Learning, specifically improving the ability for deep learning models to estimate its uncertainty and detect OOD inputs. The ability of estimate uncertainty can have positive social impact, such as detecting the hallucinations or plain wrong answers generated by large language models (LLMs). This paper is limited to smaller-scale problems like image classification on the ImageNet dataset. We will explore the application of similar techniques on large-scale LLMs in the future.

% \newpage

\bibliography{splitensemble}
\bibliographystyle{icml2024}

%%%%%%%%%%%%%%%%%%%%%%%%%%%%%%%%%%%%%%%%%%%%%%%%%%%%%%%%%%%%%%%%%%%%%%%%%%%%%%%
%%%%%%%%%%%%%%%%%%%%%%%%%%%%%%%%%%%%%%%%%%%%%%%%%%%%%%%%%%%%%%%%%%%%%%%%%%%%%%%
% APPENDIX
%%%%%%%%%%%%%%%%%%%%%%%%%%%%%%%%%%%%%%%%%%%%%%%%%%%%%%%%%%%%%%%%%%%%%%%%%%%%%%%
%%%%%%%%%%%%%%%%%%%%%%%%%%%%%%%%%%%%%%%%%%%%%%%%%%%%%%%%%%%%%%%%%%%%%%%%%%%%%%%
\newpage
\appendix
\onecolumn

\section{Experimental Setup}
\label{sec:implemen}

\paragraph{Datasets and metrics.}

We perform classification tasks on four popular image classification benchmarks, including CIFAR-10, CIFAR-100~\citep{Krizhevsky09learningmultiple}, Tiny ImageNet~\citep{deng2009imagenet} and ImageNet~\citep{krizhevsky2012imagenet} datasets. \revise{Additionally, we examine our method on the challenging long-tailed dataset, CIFAR10-LT and  CIFAR100-LT datasets~\citep{cao2019learning}.}
\begin{itemize}

\item \textbf{CIFAR10} is a collection of 60,000 32x32 color images spanning 10 different classes, such as automobiles, birds, and ships, with each class containing 6,000 images. It is commonly used in machine learning and computer vision tasks for object recognition, serving as a benchmark to evaluate the performance of various algorithms.

\item \textbf{CIFAR100} is a diverse and challenging image dataset consisting of 60,000 32x32 color images spread across 100 different classes. Each class represents a distinct object or scene, making it a comprehensive resource for fine-grained image classification and multi-class tasks. CIFAR-100 is widely used in machine learning research to evaluate the performance of models in handling a wide range of object recognition challenges.

\item \textbf{Tiny ImageNet} is a compact but diverse dataset containing thousands of small-sized images, each belonging to one of 200 categories. This dataset serves as a valuable resource for tasks like image classification, with each image encapsulating a rich variety of objects, animals, and scenes, making it ideal for training and evaluating machine learning models.

\item \textbf{ImageNet} is a widely used benchmark in the field of computer vision and machine learning. It contains approximately 1,000 images each from 1,000 different categories, totalling around 1 million images. 

\revise{\item \textbf{CIFAR10-LT \& CIFAR100-LT} are the long-tailed version of CIFAR10 and CIFAR100 datasets with imbalance ratio $\rho = 100$.}
  
\end{itemize}

For the out-of-distribution detection task, we use CIFAR-10, and CIFAR-100, Tiny ImageNet as in-distribution datasets, and use CIFAR-10, CIFAR-100, Tiny ImageNet, SVHN, LSUN, Gaussian Noise, Uniform Noise, as out-of-distribution datasets. \revise{Additionally, we adopt a more challenging OOD detection benchmark, named semantically coherent out-of-distribution detection (SC-OOD)~\citep{yang2021scood}.}

\begin{itemize}
 
\item \textbf{SVHN.}
The Street View House Numbers (SVHN) dataset is a comprehensive collection of house numbers captured from Google Street View images. It consists of over 600,000 images of house numbers from real-world scenes, making it a critical resource for tasks like digit recognition and localization. SVHN's diversity in backgrounds, fonts, and lighting conditions makes it a challenging but vital dataset for training and evaluating machine learning algorithms in the domain of computer vision.

\item \textbf{LSUN.}
The LSUN (Large-scale Scene Understanding) dataset is a vast collection of high-resolution images, primarily focused on scenes and environments. It encompasses a diverse range of scenes, including bedrooms, kitchens, living rooms, and more. LSUN serves as a valuable resource for tasks such as scene recognition and understanding due to its extensive coverage of real-world contexts and rich visual content.

\item \textbf{Gaussian
Noise and Uniform Noise.}
After introducing Gaussian Noise or Uniform Noise to the dataset, we obtain a modified dataset, which is then utilized as an OOD dataset for our experiments. We implement this operation using the library from \cite{kirchheim2022pytorch}.

\revise{\item \textbf{SC-OOD benchmark.}
The SC-OOD (Semantically Coherent Out-of-Distribution) benchmark is designed for evaluating out-of-distribution detection models by focusing on semantic coherence. This benchmark addresses the limitations of traditional benchmarks that often require models to distinguish between objects with similar semantics from different datasets, such as CIFAR dogs and ImageNet dogs.}

\end{itemize}

\revise{We use five key metrics to evaluate the performance of ID classification and OOD detection tasks.}

\begin{itemize}
    \item \revise{\textbf{Accuracy.} This is defined as the ratio of the number of correct predictions to the total number of predictions made. We report top-1 classification accuracy on the test(val) sets of ID datasets.}

    \item \revise{\textbf{FPR (95\% TPR).} This metric stands for 'False Positive Rate at 95\% True Positive Rate'. It measures the proportion of negative instances that are incorrectly classified as positive when the true positive rate is 95\%. A lower FPR at 95\% TPR is desirable as it indicates fewer false alarms while maintaining a high rate of correctly identified true positives.}

    \item \revise{\textbf{Detection Error (95\% TPR).} Detection Error at 95\% TPR is a metric that quantifies the overall error rate when the model achieves a true positive rate of 95\%. It combines false negatives and false positives to provide a single measure of error. Lower detection error values indicate better performance, as the model successfully identifies more true positives with fewer errors.}

    \item \revise{\textbf{AUROC} is short for Area Under the Receiver Operating Characteristic Curve(AUROC). This metric measures the ability of a model to distinguish between in-distribution and OOD samples. The ROC curve plots the true positive rate against the false positive rate at various threshold settings. The AUROC is the area under this curve, with higher values (closer to 1.0) indicating better discrimination between in-distribution and OOD samples.}

    \item \revise{\textbf{AURP} is short for Area Under the Precision-Recall Curve (AUPR), this metric is particularly useful in scenarios where there is a class imbalance (a significant difference in the number of in-distribution and OOD samples). It plots precision (the proportion of true positives among positive predictions) against recall (the proportion of true positives identified). Higher AUPR values suggest better model performance, especially in terms of handling the balance between precision and recall.}
    
\end{itemize}

\paragraph{Implementation Details.}

Our Split Ensemble model was trained over 200 epochs using a single NVIDIA A100 GPU with 80GB of memory, for experiments involving CIFAR-10, CIFAR-100, and Tiny ImageNet datasets. \revise{For the larger-scale ImageNet dataset, we employ 8 NVIDIA A100 GPUs, each with 80GB memory, to handle the increased computational demands.} We use an SGD optimizer with a momentum of 0.9 and weight decay of 0.0005. We also adopt a 200-epoch cosine learning rate schedule with 10 warm-up epochs and a batchsize of 256. 
Our experiments typically run for approximately 2 hours on both CIFAR-10 and CIFAR-100 datasets, \revise{whereas on the Tiny ImageNet and ImageNet datasets, they take approximately 10 hours and 24 hours, respectively.} We employ data augmentation techniques such as rotation and flip during the training phase, while the testing phase does not involve data augmentation. As for the backbone models in our experiments, we utilize the standard ResNet-18 and ResNet-34 architectures.
\revise{We heuristically decide the number of submodels in the Split-Ensemble via ablation study, where we find 8 submodels for ImageNet-1K and 5 submodels for other datasets leads to the best performance in both ID and OOD detection. The classes are grouped based on semantic similarity into subtasks for the submodels to learn.} 
For OOD detection score computation, we use the max softmax probability for the single model, max softmax probability of the mean logits for the naive ensemble, and use the probability score proposed in Section~\ref{ssec:ens_train} for our Split-Ensemble. A single threshold is used to detect OOD with score lower than the threshold. %moved from Sec5.1

% \clearpage

\section{Pseudo Code of Split-Ensemble Training}
\label{ap:alg}

The Pseudo code of Split-Ensemble training is available in Algorithm~\ref{alg}.

\begin{algorithm}
\footnotesize
	\caption{Training the Split-Ensemble model} 
	\label{alg}
	\begin{algorithmic}[1]
	    \STATE \texttt{\# Initialization and preparation}
		\STATE Load dataset $\{X,Y\}$
		\STATE Subtask label conversion $Y \rightarrow \hat{Y}_i$ as Equation~(\ref{equ:OE_label})
        \STATE Initialize Split-Ensemble $F$ with all submodels $f_i$ sharing backbone model
	    \STATE \texttt{\# Split ensemble training}
		\WHILE {Training}
            \STATE Update $F$ to minimize $\mathcal{L}_{ens}$ in Equation~(\ref{equ:ENS_loss}) with SGD
            \STATE \texttt{\# Iterative splitting and pruning}
            \IF {Epoch \% Prune\_Interval == 0}
                \STATE \texttt{\# Splitting}
                \IF {$\exists$ branch in $F$ with multiple submodel $f_i$ sharing all layers}
                    \FOR {Layers in the branch shared only by $f_i$}
                        \STATE Compute sensitivity map following Equation~(\ref{equ:SNIP}) for each $f_i$
                        \STATE Compute MCT of the layer following Equation~(\ref{equ:MCT})
                        \IF {MCT $<$ threshold}
                            \STATE Split branch at the layer
                            \STATE Break
                        \ENDIF
                    \ENDFOR
                \ENDIF
                \STATE \texttt{\# Pruning}
                \IF {FLOPs $>$ target}
                    \FOR {All submodels $f_i$}
                        \STATE Compute $\mathcal{I}^i_{\mathcal{S}}$ for all $\mathcal{S}$ following Equation~(\ref{equ:imp})
                        \STATE Rank $\mathcal{I}^i_{\mathcal{S}}$ to decide prunable structures with $\min_{\mathcal{S}}\mathcal{I}^L_{\mathcal{S}}$
                        \STATE Remove structures prunable for (all) corresponding submodels
                    \ENDFOR
                \ENDIF
            \ENDIF
		\ENDWHILE
	\end{algorithmic} 
\end{algorithm}

\section{Additional Results and Visualizations}
\label{ap:visualize}

In this section, we provide additional results in comparison with baseline methods in different settings as well as ablation results on our design choices following the discussion in Section~\ref{ssec:ablation}.

\begin{figure}[h]
    \centering
    \includegraphics[width=0.4\linewidth, keepaspectratio=true]{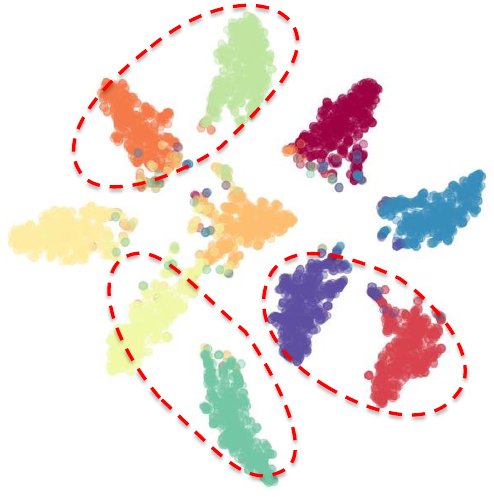}
     \vspace{-0.2cm}
    %\vspace{-0.7cm}
    \caption{\textbf{TSNE visualization of learned features from ResNet-18 with CIFAR10 test set.} Samples with different class labels are marked in different colors. The adjacent features within the dashed lines are to be grouped into one subtask.}
    \label{fig:vis_tsne}
\end{figure}

\paragraph{Subtask grouping strategy}
\label{sec:subtask-grouping}
In Section~\ref{ssec:task_split}, we propose to use the group of classes that are semantically-close to form each subtasks of the complementary task splitting. We illustrate this concept in Figure~\ref{fig:vis_tsne}, showing the natural semantic similarity we utilized to group our subtasks. Furthermore, we verify this intuition against have random assignment of classes to each subtask. As illustrated in Table~\ref{tab:ablate_group}, having semantically-close subtask grouping significantly improves the OOD detection ability of the Split-Ensemble model over that of random grouping. This improvement is more significant with more subtask splittings. We believe that semantic grouping of subtasks help the submodels to better learn the difference between ID classes and OOD classes of the subtask, as the semantically-close ID classes may share more distinct features comparing to other classes. 

\begin{table}[h]
 \caption{\textbf{Ablation on subtask grouping strategy.} Models are trained on CIFAR-100. OOD detection is against the CIFAR-10 dataset.}
  \label{tab:ablate_group}
  \vspace{0.1cm}
  \centering
  \small
  \begin{tabular}{c|c|cc} 
    \toprule
    \# splits & Subtask grouping & Accuracy & AUROC \\
    \midrule
     \multirow{2}{0.01\textwidth}{2} & Random & 77.3 & 78.9 \\
     & Semantic & \textbf{77.8} & \textbf{79.6} \\
     \midrule
     \multirow{2}{0.01\textwidth}{4} & Random  & 77.3 & 77.5 \\ 
     &  Semantic  & \textbf{77.5} & \textbf{79.1} \\
     \midrule
     \multirow{2}{0.01\textwidth}{5} & Random & 77.4 & 77.3 \\ 
     & Semantic & \textbf{77.9} & \textbf{78.9} \\
    \bottomrule
  \end{tabular}
  \vspace{-10pt}
\end{table}

\paragraph{Number of subtask splittings} 

We conducted an analysis to explore the impact of the number of splits on the accuracy and OOD detection performance of the Split-Ensemble model. Unlike traditional ensemble that repeatedly learn the same task with more submodels, Split-Ensemble always learns a complementary subtask splitting corresponding to the original task. Increasing the amount of splits will therefore enable each submodel to learn a simpler subtasks with less ID classes, intuitively leading to a model architecture with more yet smaller branches. As shown in Table~\ref{tab:ablate_split}, the Split-Ensemble accuracy is not sensitive to the number of splits, showing the scalability of our learning algorithm. For OOD detection, a larger number of splits enables each submodel to learn its OOD-aware objective more easily, therefore leading to better AUROC. Yet the performance may suffer from aggressive pruning with too much branches in the Split-Ensemble, as observed with a large MCT threshold in Table~\ref{tab:ablate_mct}. An interesting future direction would be automatically design the amount of subtask splitting and the grouping of each subtask during the training process to better fit the subtasks to the Split-Ensemble architecture.

\begin{table}[h]
 \caption{\textbf{Ablation on number of splits.} Models are trained on CIFAR-100. OOD detection is against the CIFAR-10 dataset. All models are constrained with single-model computation cost.}
  \label{tab:ablate_split}
  \vspace{0.1cm}
  \centering
  \small
  \begin{tabular}{c|ccccc} 
    \toprule
    \# splits & 2 & 4 & 5 & 8 & 10 \\
    \midrule
     Accuracy & 77.7 & \textbf{78.0} & 77.9 & 77.5 & 77.3 \\
     AUROC & 78.1 & 78.2 & 79.9 & \textbf{80.4} & 77.3 \\
    \bottomrule
  \end{tabular}
\end{table}

\begin{table}[ht]
 \caption{\textbf{Comparison between previous state-of-the-art single-model-based methods and ours on the SC-OOD CIFAR10 benchmarks.}  The results are reported for models with ResNet-18 backbone. Best score in \textbf{bold}, second best \underline{underlined}.}
  \label{tab:single-cifar10-SCOOD}
  \vspace{0.1cm}
  \centering
  \small
  \begin{tabular}{l|c|ccc} 
    \toprule
    Method                & Additional Data & FPR95 $\downarrow$ & AUROC $\uparrow$ & AUPR $\uparrow$ \\
    \midrule
    ODIN                  &      \faTimes           & 52.0                    & 82.0                    & 85.1                   \\
EBO                   &        \faTimes         & 50.0                    & 83.8                    & 85.1                   \\
OE                    &     \faCheck            & 50.5                    & 88.9                    & 87.8                   \\
MCD                   &       \faCheck          & 73.0                    & 83.9                    & 80.5                   \\
UDG                   &       \faTimes          & 55.6                    & 90.7                    & 88.3                   \\
UDG                &      \faCheck           & \textbf{36.2}           & \textbf{93.8}           & \textbf{92.6}          \\
Split-Ensemble (ours) &      \faTimes           & \underline{45.5}              & \underline{91.1}              & \underline{89.9}           \\
    \bottomrule
  \end{tabular}
\end{table}

\revise{\paragraph{Additional OOD detection results on CIFAR10 with SC-OOD benchmark}
\label{sec:cifar10-SCOOD}
We further compare our methods with previous state-of-the-art methods. In Table.~\ref{tab:single-cifar10-SCOOD}, our Split-Ensemble model outperforms single-model approaches in OOD detection without incurring additional computational costs or requiring extra training data. Its consistent high performance across key metrics highlights its robustness and efficiency, underscoring its practical utility in OOD tasks. In Table.~\ref{tab:ensemble-cifar10-SCOOD}, our Split-Ensemble model consistently outshines other ensemble-based methods in both image classification and OOD detection, achieving top rankings across all key metrics, which underscores the model's efficiency and effectiveness.
}

\begin{table}[h]
 \caption{\textbf{Comparison between previous state-of-the-art ensemble-based methods and ours on the SC-OOD CIFAR10 benchmarks.}  The results are reported for models with ResNet-18 backbone. Best score in \textbf{bold}, second best \underline{underlined}.}
  \label{tab:ensemble-cifar10-SCOOD}
  \vspace{0.1cm}
  \centering
  \small
  \begin{tabular}{l|c|ccc} 
    \toprule
    Method                & FLOPs & FPR95 $\downarrow$ & AUROC $\uparrow$ & AUPR $\uparrow$ \\
    \midrule
Naive Ensemble        & 4x    & \underline{42.3}                    & 90.4                    & \underline{90.6}                   \\
MC-Dropout         & 4x    & 54.9                    & 88.7                    & 88.0                   \\
MIMO                  & 4x    & 73.7                    & 83.5                    & 80.9                   \\
MaskEnsemble          & 4x    & 53.2                    & 87.7                    & 87.9                   \\
BatchEnsemble         & 4x    & 50.4                    & 89.2                    & 88.6                   \\
FilmEnsemble          & 4x    & \textbf{42.6}           & \textbf{91.5}           & \textbf{91.3}          \\ \midrule
Split-Ensemble (ours) & 1x    & 45.5                    & \underline{91.1}                    & 89.9                  \\
    \bottomrule
  \end{tabular}
\end{table}

\paragraph{Robustness to corruption}
We evaluated our method and ensemble baselines on the corrupted dataset, CIFAR-10-C. In~\cref{tab:rob}, We follow~\citet{ovadia2019can} to report averaged top-1 accuracy (\%) and ECE metric (\%) across different corruption types for the fixed corruption severity level, which has a 1 to 5 range. It can be seen that even under data corruptions, our method has either the best or the second best in accuracy while being consistently well-calibrated and low-complexity among the ensemble baselines. 

\begin{table}[h]
  \caption{\textbf{Accuracy ($\uparrow$)/expected calibration error ($\downarrow$) on the corrupted CIFAR-10-C dataset}. Best score for each metric in \textbf{bold}, second-best \underline{underlined}. We implement all baselines using default hyperparameters. All accuracies/ECEs are given in percentage with ResNet-18 as backbone.}
  \vspace{0.1cm}
  \label{tab:rob}
  \centering
  \small
  %\resizebox{.6\linewidth}{!}{
  %\setlength\tabcolsep{4pt}
  \begin{tabular}{l|c|cccccccc}
    \toprule
    %\multicolumn{2}{c}{Part}                   \\
   % \cmidrule(r){1-2}
    Method    & FLOPs & Clean & Severity 1 & Severity 2 & Severity 3 & Severity 4 & Severity 5 \\
    \midrule
    Naive Ensemble & 4x    & \textbf{95.7} / 3.9 & \textbf{91.4} / 7.7 & \textbf{85.1} / 13.6 & \underline{78.6} / 19.7 & \underline{71.9} / 26.0 & 59.4 / 37.7 \\ 
    \midrule
    MC-Dropout    & 4x     & 92.4 / 3.5 & 88.3 / \underline{5.7} & 82.7 / \underline{9.5}  & 77.0 / \underline{13.6} & 70.9 / 18.2 & \underline{59.6} / 26.2 \\ 
    MIMO          & 4x     & 84.4 / \textbf{1.1} & 79.0 / \textbf{3.4} & 72.0 / \textbf{6.9 } & 65.7 / \textbf{10.8} & 59.2 / \textbf{15.1} & 49.8 / \textbf{21.4} \\ 
    MaskEnsemble   & 4x    & 94.2 / 3.3 & 89.3 / 6.9 & 83.6 / 10.9 & 77.6 / 15.4 & 70.9 / 20.2 & 58.9 / 29.9 \\ 
    BatchEnsemble  & 4x    & 93.8 / 3.1 & 88.8 / 5.7 & 82.7 / 10.1 & 76.1 / 15.4 & 69.4 / 20.2 & 57.6 / 29.1 \\ 
    FilmEnsemble   & 4x    & 93.8 / \underline{3.0} & 89.1 / 6.5 & 84.1 / 10.5 & 77.7 / 13.9 & 71.5 / 20.2 & 59.2 / 27.2 \\ 
   \midrule
    \textbf{Split-Ensemble (Ours)} & 1x     & \underline{95.5} / 4.1 & \underline{90.7} / 7.0 & \underline{84.8} / 10.5 & \textbf{79.2} / 13.7 & \textbf{73.2} / \underline{16.7} & \textbf{61.7} / \underline{22.7} \\ 
    \bottomrule 
  \end{tabular}
  %}
\end{table}

\begin{figure}[h]
    \centering
    \includegraphics[width=0.9\linewidth, keepaspectratio=true]{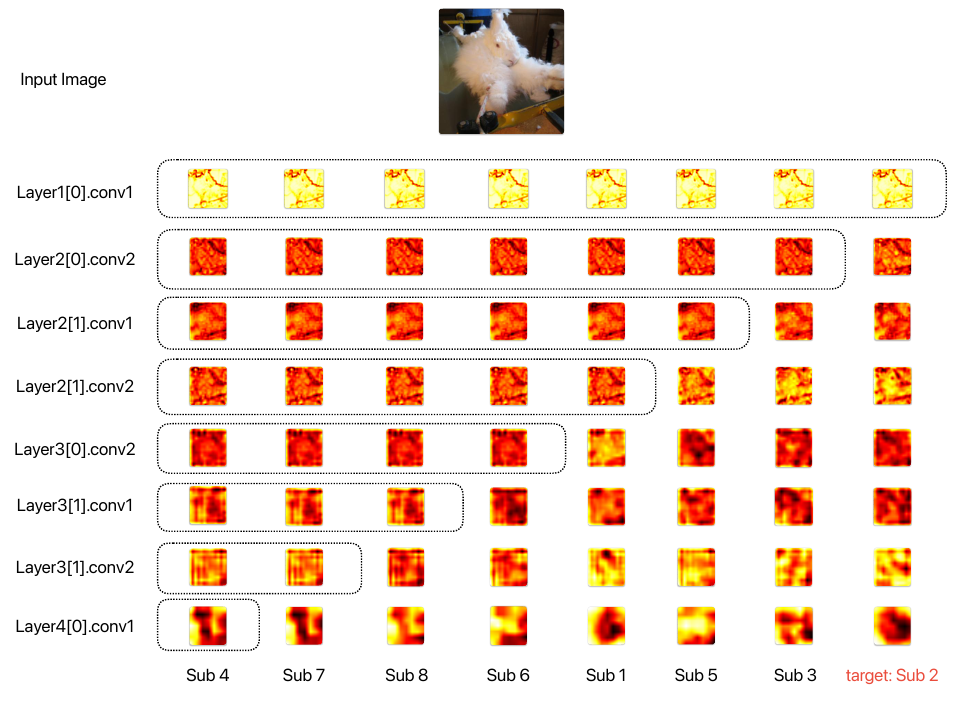}
     \vspace{-0.2cm}
    %\vspace{-0.7cm}
    \caption{\textbf{Visualization of Split-Ensemble's learned features using Score-CAM.} The number of splits is set to 8 and the model is trained on ImageNet1K with ResNet-18 as backbone. The feature maps within the dashed lines across the layers indicate shared representations. The input image's class is 'Angora', targeted by submodel 2.}
    \label{fig:vis_feat}
\end{figure}

\revise{
\paragraph{Model activation map visualization}
We visualize the learned feature map activations of a Split-Ensemble model across different layers using Score-CAM~\citep{wang2020score} in Figure~\ref{fig:vis_feat}. The shared feature maps, delineated by dashed lines, represent the common features extracted across different submodels, emphasizing the model's capacity to identify and leverage shared representations. The distinct feature maps outside the dashed boundaries correspond to specialized features pertinent to individual sub-tasks, demonstrating the Split-Ensemble model's ability to focus on unique aspects of the data when necessary. This visualization underscores the effectiveness of the Split-Ensemble architecture, highlighting its dual strength in capturing both shared and task-specific features within a single, cohesive framework, thereby bolstering its robustness and adaptability in handling diverse image classification and OOD detection tasks.
}

\begin{figure}[h]
    \centering
    \includegraphics[width=0.9\linewidth, keepaspectratio=true]{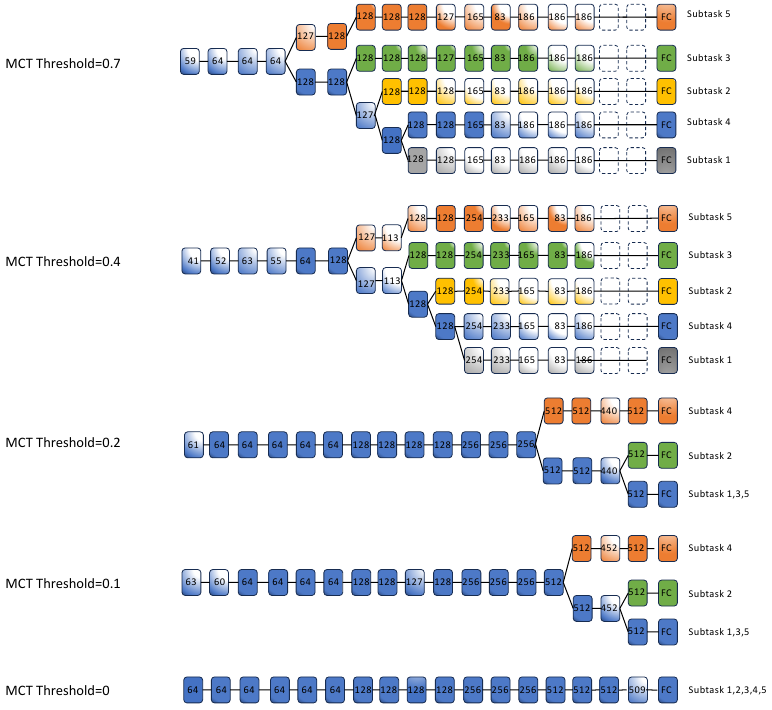}
    \vspace{-0.2cm}
    \caption{\textbf{Visualization of Split-Ensemble architectures under different MCT threshold.} The number of splits is set to 5. Number in each block denotes the number of filters in the layer.}
    \label{fig:vis_arch}
\end{figure}

\paragraph{Model architecture visualization}
We visualize the Split-Ensemble models achieved under different MCT thresholds in Figure~\ref{fig:vis_arch}, as discussed in Table~\ref{tab:ablate_mct} of Section~\ref{ssec:ablation}. Models here use 5 splits and are trained on CIFAR-100 dataset with ResNet-18 as backbone. With a larger MCT threshold, the model will split into more branches at earlier layers. Meanwhile the model will also be pruned more aggressively to keep the overall computation cost unchanged. We can clearly see that with a proper MCT threshold, our method can learn a tree-like Split-Ensemble architecture with different submodels branching out at different layers, as designed by our iterative splitting and pruning algorithm in Section~\ref{sec:prune}. 

\end{document}